\documentclass[sigconf]{acmart}

\usepackage{multirow}
\usepackage{caption}
\usepackage{tablefootnote}
\usepackage{lineno}
\usepackage{graphicx}
\usepackage{amsmath}
\usepackage{subfigure}
\usepackage{hyperref}
\usepackage{xcolor}
\usepackage[linesnumbered,ruled,vlined]{algorithm2e}

\AtBeginDocument{%
  }

\copyrightyear{2023} 
\acmYear{2023} 
\setcopyright{acmlicensed}\acmConference[MM '23]{Proceedings of the 31st ACM International Conference on Multimedia}{October 29-November 3, 2023}{Ottawa, ON, Canada}
\acmBooktitle{Proceedings of the 31st ACM International Conference on Multimedia (MM '23), October 29-November 3, 2023, Ottawa, ON, Canada}
\acmPrice{15.00}
\acmDOI{10.1145/3581783.3612378}
\acmISBN{979-8-4007-0108-5/23/10}

\begin{document}

\title{Underwater Image Enhancement by Transformer-based Diffusion Model with Non-uniform Sampling for Skip Strategy}

\author{Yi Tang}
\orcid{0000-0002-4882-5234}
\affiliation{%
  \institution{Kitami Institute of Technology}
  \city{Kitami}
  \state{Hokkaido}
  \country{Japan}
}
\affiliation{%
	\institution{Kyushu University}
	\city{Fukuoka}
	\state{Fukuoka}
	\country{Japan}
}
\email{tangyi@mail.kitami-it.ac.jp}

\author{Hiroshi Kawasaki}
\affiliation{%
	\institution{Kyushu University}
	\city{Fukuoka}
	\state{Fukuoka}
	\country{Japan}
}
\email{kawasaki@ait.kyushu-u.ac.jp}

\author{Takafumi Iwaguchi}
\affiliation{%
	\institution{Kyushu University}
	\city{Fukuoka}
	\state{Fukuoka}
	\country{Japan}
}
\email{iwaguchi@ait.kyushu-u.ac.jp}

\renewcommand{\shortauthors}{Yi Tang et al.}

\begin{abstract}
  In this paper, we present an approach to image enhancement with diffusion model in underwater scenes. Our method adapts conditional denoising diffusion probabilistic models to generate the corresponding enhanced images by using the underwater images and the Gaussian noise as the inputs. Additionally, in order to improve the efficiency of the reverse process in the diffusion model, we adopt two different ways. We firstly propose a lightweight transformer-based denoising network, which can effectively promote the time of network forward per iteration. On the other hand, we introduce a skip sampling strategy to reduce the number of iterations. Besides, based on the skip sampling strategy, we propose two different non-uniform sampling methods for the sequence of the time step, namely piecewise sampling and searching with the evolutionary algorithm. Both of them are effective and can further improve performance by using the same steps against the previous uniform sampling. In the end, we conduct a relative evaluation of the widely used underwater enhancement datasets between the recent state-of-the-art methods and the proposed approach. The experimental results prove that our approach can achieve both competitive performance and high efficiency. Our code is available at \href{mailto:https://github.com/piggy2009/DM_underwater}{\color{blue}{https://github.com/piggy2009/DM\_underwater}}.
\end{abstract}




\maketitle

\section{Introduction}
Image enhancement is a fundamental technique for image processing, whose purpose is to improve the quality of image content and recover the original pixel-wise information from low-quality images, such as low-light enhancement \cite{zhang2019kindling,guo2021dynamic}, image dehaze \cite{liu2019learning} and deraining \cite{yang2020single}. Recently, remotely operated underwater vehicles (ROV) and related research for marine environments, such as exploring marine life and protecting ecosystems \cite{kimball2018artemis}, have become popular. However, the quality of captured images/videos by the ROV is usually too low to be further analyzed by the high-level application, such as object tracking, detection or scene recognition. Accordingly, underwater image enhancement (UIE) has become an important technology to be deployed into these exploration systems. However, compared with the other enhancement tasks, underwater scenarios are complex and diverse, such as color distortion, low and biased illumination, and heavy and non-uniform blurriness~\cite{peng2021u}. Therefore, UIE usually suffers from different types of image noise against the single noise in other tasks, such as low-light enhancement and super-resolution. 

Early approaches~\cite{chiang2011underwater,drews2016underwater,li2016underwater} mainly rely on a physical model, such as Retinex model~\cite{land1977retinex} or underwater degradation imaging model~\cite{mcglamery1980computer}. These methods are able to cope with some simple underwater scenes like shallow water areas, however, for more complex scenes, such as heavy blurriness or low illumination scenes, it is difficult to be used because a single physical model is not appropriate to explain the real degradation. To solve the problem, recent approaches introduce deep learning-based models~\cite{li2019underwater,fabbri2018enhancing,islam2020simultaneous}, where a large-scale paired data, namely underwater image and its corresponding ground truth, is used to train a neural network to obtain enhanced images. Since it is almost impossible to retrieve a pair of real data for the underwater scenario, the synthetic images created by Li {\it et al.}~\cite{li2019underwater} are commonly used for training and testing. Previous approaches are to exploit the generative adversarial network (GAN) without using paired data. With those techniques, unpaired data~\cite{li2018emerging} 
or simulated underwater images from in-air images~\cite{li2017watergan} are used. Recently, a new multi-scale dense GAN for enhancing underwater images is proposed~\cite{guo2019underwater}. However, due to the existing two adversarial networks, GAN-based techniques are known to be unstable and frequently introduce model collapse. In this paper, we introduce a new generation model, called the diffusion model (DM), which is more stable and achieves faster training than GAN. 

The diffusion model consists of the forward process and the reverse process. The forward process is a Markov process, whose function is to corrupt the image by progressively appending the normal Gaussian noise. Image details are removed until it becomes pure noise. During the reverse process, a neural network is trained to gradually remove the noise from the forward process by giving the pure noise. Originally, since the noise image given in the reverse process is random, the generated images have huge diversity, which is contradict to our purpose to generate a specific image corresponding to the input underwater image. Hence, we employ a denoising network with an extra image as the condition to provide the image content that we want to restore. Additionally, it is well known that the inference time of the diffusion model is relatively long because the reverse process is an iterative process for several thousand iterations to generate a clean image.
%is generated by over thousands of network forward. 
In this paper, we address this issue in two aspects. First, we propose a lightweight transformer-based denoising network to decrease the runtime of the network forward during an iteration. Second, the skip sampling strategy \cite{song2020denoising} is introduced to reduce the number of the iteration. Moreover, considering the inflexibility of the original uniform sampling strategy, we propose two non-uniform sampling strategies. The first one is a piecewise sampling approach and the other is a search method based on the evolutionary algorithm, which simulates the mechanisms of biological evolution like mutation and recombination to select the best candidate. It is experimentally shown that both of the proposed sampling strategies can retain the original efficiency and achieve better performance than the original uniform sampling strategy.

In summary, the main contribution of this paper is as follows:
\begin{itemize}
	\item We introduce a conditional diffusion framework for an underwater image enhancement, which can generate specific enhanced images and achieve competitive performance in widely known techniques.
	%used underwater datasets. 
	
	\item We propose a lightweight transformer-based neural network as the denoising network in the diffusion framework, which can effectively improve the quality of images and decrease the runtime in the denoising process.

	\item For the skip sampling strategy in the reverse process, we propose two different ways to sample the time steps. Both of them can retain the original number of iterations and further improve the enhancement performance. 
\end{itemize}

\section{Related works}

\subsection{Underwater image enhancement}

The recent methods in the community of underwater image enhancement can be briefly divided into two categories, namely bottom-up methods and top-down methods. 

The bottom-up methods are heuristic frameworks. This kind of method often relies on the standard physical model Retinex model~\cite{land1977retinex} and underwater optical imaging model~\cite{mcglamery1980computer} to improve the quality of underwater images. For example, in~\cite{he2010single} and~\cite{chiang2011underwater}, they introduce the dark channel prior (DCP) to estimate the atmospheric light and transmission maps in the optical imaging model so that the clear images can be accurately predicted by this model. Previous method~\cite{fu2014retinex} exploits the Retinex model to decompose the reflectance and the illumination, thus proposing an effective color correction strategy to address the color distortion. Besides, dynamic pixel range stretching~\cite{iqbal2010enhancing}, fusion model~\cite{ancuti2012enhancing} and pixel distribution adjustment~\cite{ghani2015underwater} are proposed in underwater enhancement tasks. These approaches are effective to some extent. However, due to the diverse marine scenarios and variable illumination by the depth of the sea, the robustness of these approaches cannot satisfy these requirements. Moreover, these methods usually suffer from heavy computation, so most of them are low efficiency.   

The other category is top-down methods, namely learning-based frameworks. Recently, deep learning-based methods achieve competitive performance in this community. In the beginning, due to the shortage of paired training images, a variety of methods introduce generative adversarial networks for underwater image enhancement, such as WaterGAN~\cite{li2017watergan}, FUnIE~\cite{islam2020fast}, UGAN~\cite{fabbri2018enhancing} and UIE-DAL~\cite{uplavikar2019all}. After the release of new underwater datasets like UIEB~\cite{li2019underwater} and LSUI~\cite{peng2021u}, some complex frameworks, such as WaterNet~\cite{li2019underwater}, Ucolor~\cite{li2021underwater}, TSDA~\cite{jiang2022two} and Ushape\cite{peng2021u}, are proposed and achieve the-state-of-the-art performance. For example, Li et al~\cite{li2021underwater} propose a transmission map guided network, which combines medium transmission map and multi-color inputs and allows the network to learn robust features for enhancement. PUIENet~\cite{Fu_2022} combines the conditional variational autoencoder with adaptive instance normalization and proposes an adjustable enhancement approach. Ushape\cite{peng2021u} proposes a Transformer-based method, which introduces two kinds of transformer structures to improve image quality. RCTNet~\cite{kim2021representative} introduces attention mechanisms and color transformation to address the issues in underwater scenes. 

\begin{figure*}
	\centering
	\subfigure[]{\includegraphics[width=1.0\textwidth]{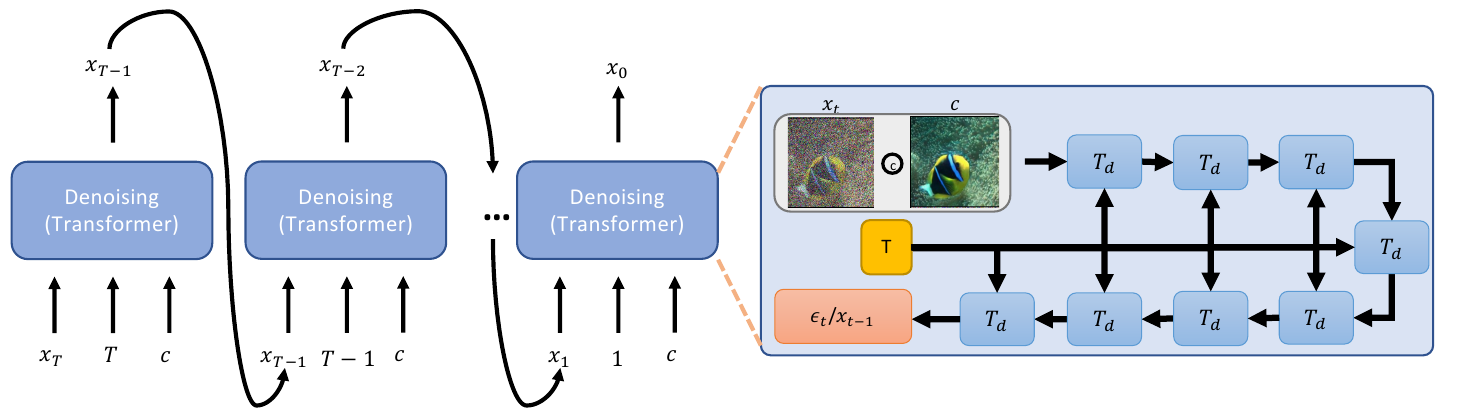}} 
	\subfigure[]{\includegraphics[width=0.45\textwidth]{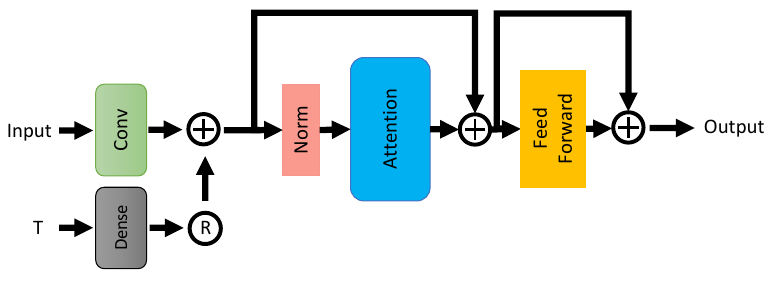}} 
	\hfill
	\subfigure[]{\includegraphics[width=0.45\textwidth]{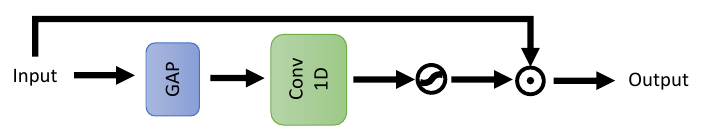}}
	\caption{The proposed framework and specific neural network. (a) The iterative refinement of the diffusion model. The model is fed by the noisy image $\mathbf{x}_T$, conditional image $c$ and time step $t$ to generate the clear image step by step. (b) The transformer block $T_d$. The denoising network consists of these blocks, which are used to encode and refine the features. (c) The channel-wise attention module. We introduce a lightweight attention module, which is able to learn robust feature representatives and decrease the scale of parameters.}
	\label{framework}
\end{figure*}

\subsection{Diffusion model}

Following the GAN, Variational Autoencoder (VAE) and Flow-based models, the denoising diffusion probabilistic model (DDPM)~\cite{ho2020denoising} is a new generative framework. It contains two processes, namely the forward diffusion process and the reverse diffusion process. The former process is a Markov chain, in which Gaussian noises are gradually appended into the input data so that we can obtain the latent variables $x_t$ by sequential sampling, where $t = \{1, 2, ..., T\}$. For each step in the forward process, we can represent it as a Gaussian transition $q(\mathbf{x}_t \vert \mathbf{x}_{t-1}) = \mathcal{N}(\mathbf{x}_t; \sqrt{1 - \beta_t} \mathbf{x}_{t-1}, \beta_t\mathbf{I})$, where $\{\beta_t \in (0, 1)\}_{t=1}^T$ is a variance schedule. According to the Markov chain and iterative relation, we can sample $x_t$ at any arbitrary time step $t$:
\begin{equation}
	\label{forward}
	\begin{aligned}
		x_t &= \sqrt{\alpha_t} x_0 + \sqrt{1-\alpha_t} \boldsymbol{\epsilon}
	\end{aligned}
\end{equation} where $\alpha_t = \prod_{i=1}^t (1 - \beta_i)$, $\boldsymbol{\epsilon} \sim  \mathcal{N}(\mathbf{0}, \mathbf{I})$. 

For the reverse diffusion process $q(\mathbf{x}_{t-1} \vert \mathbf{x}_t)$, another Gaussian transition $p_{\theta}(\mathbf{x}_{t-1} \vert \mathbf{x}_{t}) = \mathcal{N}(\mathbf{x}_{t-1}; \mu_{\theta}(x_t, t), \sigma_{\theta}(x_t, t)\mathbf{I})$ is used for estimation. Among them, $\mu_{\theta}(x_t, t)$ can be denoted as the combination of $x_t$ and a predicted noise $\boldsymbol{\epsilon}_{\theta}(x_t, t)$ by a neural network. The network optimization can be addressed as follows:  
\begin{equation}
	\label{forward}
	\begin{aligned}
		\mathop{\min}_{\theta} \mathbb{E}_{x_0 \sim q(x_0), \epsilon \sim \mathcal{N}(\mathbf{0}, \mathbf{I}), t}  \|\boldsymbol{\epsilon}_t - \boldsymbol{\epsilon}_\theta(\mathbf{x}_t, t)\|^2 
	\end{aligned}
\end{equation}

After training the denoising network, the output can be sampled by the reverse process as below:

\begin{equation}
	\label{backward}
	\begin{aligned}
		x_{t-1} &= \frac{1}{\sqrt{1 - \beta_t}} \Big( \mathbf{x}_t - \frac{\beta_t}{\sqrt{1 - \bar{\alpha}_t}} \boldsymbol{\epsilon}_\theta(\mathbf{x}_t, t) \Big) + \sigma_t\boldsymbol{z}
	\end{aligned}
\end{equation} where $\boldsymbol{z} \sim \mathcal{N}(\mathbf{0}, \mathbf{I})$. We can notice that the reverse process is also an iterative process. We start from $x_T$ and generate the predicted $x_0$ in the end. 

Based on the DDPM, a variety of methods are proposed in different tasks, such as human motion domain~\cite{tevet2022human}, super-solution~\cite{rombach2022high,saharia2022image}, domain transfer~\cite{kim2022diffusionclip,rombach2022high} and so on. For example, SR3~\cite{saharia2022image} proposes a conditional diffusion model, so that we can feed a conditional image into the model, and then obtain a corresponding output. HRIS~\cite{rombach2022high} introduces an extra encoder-decoder model to compress the input and recover the output, thus speeding up the reverse process. RePaint~\cite{lugmayr2022repaint} employs an unconditional DDPM as the generative prior and proposes a novel reverse process to sample the unmasked regions.

\section{Proposed method}

\subsection{Overview}

As shown in Figure.\ref{framework}, our framework mainly contains the diffusion process, transformer-based denoising network and channel-wise attention module. Our purpose is to generate the corresponding enhanced image by giving a low-quality image. It cannot be satisfied by using the original diffusion model with indeterminate results. Therefore, we introduce the conditional diffusion model. Given the input $(x_{t}, c, t)$, where $x_t$ is a noisy image, $c$ denotes a conditional image, and $t$ represents the time step, the network is used to estimate the noisy distribution $\boldsymbol{\epsilon_t}$. During the training process, we use L1 loss to optimize the network:

\begin{equation}
	\label{forward}
	\begin{aligned}
		L_{s} = \|\boldsymbol{\epsilon}_t - \boldsymbol{\epsilon}_\theta(\mathbf{x}_t, c, t)\|
	\end{aligned}
\end{equation} where $\boldsymbol{\epsilon}_\theta$ is predicted noisy image. 

The reverse process in the diffusion model is an iterative denoising process or a reverse Markovian process, namely to estimate $q(\mathbf{x}_{t-1} \vert \mathbf{x}_t)$. We will start to estimate it with a pure Gaussian noise $\mathbf{x}_T$. That is:

\begin{equation}
	\label{backward2}
	\begin{aligned}
		p_\theta(x_{0:T} \vert c) &= p(\mathbf{x}_T) \prod^T_{t=1} p_\theta(\mathbf{x}_{t-1} \vert \mathbf{x}_t, c) \\
		p_\theta(\mathbf{x}_{t-1} \vert \mathbf{x}_t, c) &= \mathcal{N}(\mathbf{x}_{t-1}; \boldsymbol{\mu}_\theta(\mathbf{x}_t, c, t), \boldsymbol{\Sigma}_\theta(\mathbf{x}_t, c, t))
	\end{aligned}
\end{equation} When we adopt a very small $\beta_t$, the reverse process for each step can be treated as a Gaussian process. Therefore, we need to design a network to estimate the mean value $\boldsymbol{\mu}_{\theta}(\cdot)$ and variance $\boldsymbol{\Sigma}_{\theta}(\cdot)$. 

According to \cite{ho2020denoising}, the mean value can be written as:

\begin{equation}
	\label{backward5}
	\begin{aligned}
		\boldsymbol{\mu}_{\theta}(\mathbf{x}_t, c, t) =  \frac{1}{\sqrt{1 - \beta_t}} \Big( \mathbf{x}_t - \frac{\beta_t}{\sqrt{1 - \alpha_t}} \boldsymbol{\epsilon}_\theta(\mathbf{x}_t, c, t) \Big)
	\end{aligned}
\end{equation}

After that, we can sample this Gaussian distribution by using the reparameterization trick. Given a Gaussian distribution $\mathbf{x} \sim q_\phi(\mathbf{z}\vert\mathbf{y}^{(i)}) = \mathcal{N}(\mathbf{z}; \boldsymbol{\mu}, \boldsymbol{\sigma}^{2}\boldsymbol{I})$, we can sample one $\mathbf{x}$ by using $\mathbf{z} = \boldsymbol{\mu} + \boldsymbol{\sigma} \odot \boldsymbol{\epsilon} $, where $\boldsymbol{\epsilon} $ is a normal Gaussian distribution and $\odot$ refers to element-wise product. Finally, we use the equation as follows to iteratively generate the sample images until we obtain the final image $x_0$.

\begin{equation}
	\label{backward2}
	\begin{aligned}
		x_{t-1} &= \frac{1}{\sqrt{1 - \beta_t}} \Big( \mathbf{x}_t - \frac{\beta_t}{\sqrt{1 - \alpha_t}} \boldsymbol{\epsilon}_\theta(\mathbf{x}_t, c, t) \Big) + \sigma_t\boldsymbol{z}
	\end{aligned}
\end{equation} Note that compared with the original diffusion model, our denoising network $\boldsymbol{\epsilon}_\theta(\mathbf{x}_t, c, t)$ inserts a conditional input $c$. In our training process, according to Eq.\ref{forward}, $x_t$ is the noisy high-quality image by adding the Gaussian noise on the clean image $x_0$. $c$ is the corresponding underwater image, namely a low-quality image. The effect of the conditional image is to provide the context information. In other words, the conditional image is used to guide the diffusion model to generate the specific image, otherwise, the generated image is indeterminate.

%In the training phase, $x_t = \sqrt{\alpha_t} x_0 + \sqrt{1-\alpha_t} \boldsymbol{\epsilon}$, where $x_0$ is the ground truth, namely the high-quality image. $c$ is the corresponding underwater image, namely a low-quality image. So, $x_t$ is the noisy version of $x_0$, not $c$. The effect of the conditional image is to provide the context information. In other words, the conditional image is used to guide the diffusion model to generate the specific image, otherwise, the generated image is random without the conditional image.

\subsection{Trasnformer-based denoising network}

In this paper, we introduce a transformer-based network for denoising prediction. Compared with the original one, our network is able to exploit the shallower structure to predict the noisy image, thus generating the enhanced image. For the reverse process, due to a large number of iterations, the runtime is very slow. Inspired by the Transformer structure in \cite{zamir2021restormer}, we do not use the image patches as tokens to compute the weights in the self-attention module in the spatial dimension, ours is to compute the self-attention across channels. This kind of structure can reduce the computational complexity and be easier to learn the related color information to revise the color distortion in low-quality tasks. With our shallower and lightweight network, the efficiency of the entire diffusion model can be improved to some extent. The specific structure is shown in Figure.\ref{framework} (b) and (c). 

Given the inputs, namely noisy image $\mathbf{x}_t$, the conditional image $c$ and time step $t$, we firstly concatenate two images to a 6-channel image $I_t$ and then feed it into the network. After, we introduce a convolution layer to adjust the channel numbers, while the time step is fed into a fully connected (dense) layer for encoding. Then, the feature vector is reshaped into the feature maps $E_t$ and added to the feature maps from the 6-channel image. This process can be written as below:

\begin{equation}
	\label{input}
	\begin{aligned}
		I_t &= \Phi_\theta(Con(\mathbf{x}_t, c)) \\
		E_t &= R(D_\theta(t)) \\
		F &= I_t + E_t
	\end{aligned}
\end{equation} where $\Phi(\cdot)$, $Con(\cdot)$, $D(\cdot)$ and $R(\cdot)$ are convolution, concatenation, fully connected and reshape operators, respectively. $F$ denotes the intermediate feature maps. 

In order to obtain robust feature representations, we introduce transformer blocks to replace convolution blocks in the UNet. The advantage is to decrease the scale of the deep model and number of the parameters. As shown in Figure.\ref{framework} (b), we only use eight transformer blocks to complete the encoding and decoding. Originally, the training and inference time of diffusion models are slow. If we still deploy the big model, the efficiency is much lower. Therefore, considering the memory complexity and low efficiency of the conventional self-attention module~\cite{wang2020eca}, we employ the channel-wise attention, which can dramatically decrease the scale of parameters and improve efficiency to some extent. 

The feature encoding in the transformer block can be formulated as below:

\begin{equation}
	\label{NRL}
	\begin{aligned}
		\hat{F} &= Att(Ln(F)) + F \\
		\tilde{F} &= FF(Ln(\hat{F})) + \hat{F}
	\end{aligned}
\end{equation} where $F, \hat{F}, \tilde{F} \in R^{H \times W \times C}$ are input, intermediate and output feature maps. $Ln(\cdot)$ is the layer normalization. $Att(\cdot)$ denotes the self-attention module, whose specific structure is shown in Figure.\ref{framework} (c). Notice that the deployed self-attention is straightforward. There is only one convolution layer containing learnable parameters and the number of parameters is lower. Its formulation can be written as follows:

\begin{equation}
	\label{NRL}
	\begin{aligned}
		\hat{F} &= \phi_\theta(G(F)) \\
		\tilde{F} &= F + sigmoid(\hat{F})
	\end{aligned}
\end{equation} where $G(\cdot)$ represents the global average pooling, whose function is to convert the feature map $F \in R^{H \times W \times C}$ to a vector with shape $R^{1 \times C}$. $\phi_\theta(\cdot)$ is the 1D-convolution operator.

\subsection{Skip sampling strategy}

Theoretically, the value of time step $T$ should be as large as possible. However, as the inference process is an iterative process like a recurrent network, the network needs to be forwarded for $T$ times. Therefore, the large value of $T$ will gain heavy computation and cause heavy time-consuming. To improve the efficiency of the inference process, one way is to decrease the runtime of the denoising network. The other is to reduce the number of iterations. For the first way, we design a lightweight network to guarantee high speed in each iteration. Then, we employ an efficient method of the iterative implicit probabilistic model (DDIM)~\cite{song2020denoising}, whose training process is the same as DDPM, but it can speed up the inference stage by introducing an alternative non-Markovian process. Specifically, the new model can modify Eq.\ref{backward2}, so that the random term will be removed and the reverse process can exploit the skip sampling strategy. The process can be written as below:

\begin{equation}
	\label{backward_ddim}
	\begin{aligned}
		\mathbf{x}_{t-1} 
		&= \sqrt{\alpha_{t-1}}\mathbf{x}_0 + \sqrt{1 - \alpha_{t-1} - \sigma_t^2} \boldsymbol{\epsilon}_\theta(\mathbf{x}_t, c, t) + \sigma_t^2 \mathbf{z}
	\end{aligned}
\end{equation} where the $x_0$ can be rewritten by using Eq.\ref{forward} and $\sigma_t^2$ can be formulated as below:

\begin{equation}
	\label{backward_ddim2}
	\begin{aligned}
		\mathbf{x}_{0} &= \frac{\mathbf{x}_{t} - \sqrt{1 - \alpha_t}\boldsymbol{\epsilon}_\theta(\mathbf{x}_t, c, t)}{\sqrt{\alpha_t}}
	\end{aligned}
\end{equation}

\begin{equation}
	\label{backward_ddim2}
	\begin{aligned}
		\sigma_t^2 &= \tilde{\beta}_t = \frac{1 - \alpha_{t-1}}{1 - \alpha_t} \beta_t
	\end{aligned}
\end{equation}

For the implementation, let $\sigma_t^2 = \eta \cdot \tilde{\beta}_t$. Here, $\eta$ will be treated as a hyperparameter. When it is set to $0$, the third term in Eq.\ref{backward_ddim} will be removed. The random term is removed and the sampling process becomes deterministic. Therefore, the inference process can use an interval step sequence $\{\tau_1, \tau_2, ..., \tau_S \}$, where $|S| < |T|$. 

Previous methods~\cite{kim2022diffusionclip,song2020denoising} usually exploit a uniform sampling way $\mathcal{U} \sim [a, b]$ with a fixed stride $d$. It is effective but not flexible. Besides, we observe that the former part of the reverse process is more important than the latter part. Therefore, we propose a piecewise sampling method, which is to employ different sampling strides in the time step sequence. Specifically, we split the sequence into two pieces and use two different strides for sampling. The specific process can be formulated as follows:

\begin{equation}
	\label{sampling}
	\tau_{s} =
	\begin{cases}
		\tau_{0} + (s - 1) * d_1 & \tau_{s} \in [a, c], \tau_{0} = a\\
		\tau_{0} + (s - 1) * d_2 &\tau_{s} \in [c, b], \tau_{0} = c 
	\end{cases}
\end{equation} where $d_1$ and $d_2$ are different strides in different ranges ($[a, c]$ and $[c, b]$). Through the proposed sampling, we can further improve the performance of the diffusion model.

In order to explore a more reasonable sampling sequence, notice that the sampling sequence can be regarded as the gene sequences. We can use the evolutionary algorithm to disuse the bad offspring and select the better one. This entire process can be written in the Algorithm.~\ref{evolutionary}.

\begin{algorithm}[]
	\SetKwInput{KwInput}{Input}                % Set the Input
	\SetKwInput{KwOutput}{Output}              % set the Output
	\newcommand\mycommfont[1]{\footnotesize\ttfamily\textcolor{blue}{#1}}
	\SetCommentSty{mycommfont}
	\DontPrintSemicolon
	
	\KwInput{$pop$ population dictionary, $L_g$ length of genes, probability $p_c$ and $p_m$ for crossover and mutation, $epoch$ of iterations}
	\KwOutput{gene sequence with the best performance}
	%\KwData{Testing set $x$}
	
	% Set Function Names
	\SetKwFunction{FCrossover}{Crossover}
	\SetKwFunction{FMutation}{Mutation}
	\SetKwFunction{FRandom}{Random}
	\SetKwFunction{FMain}{Main}
	
	\tcc{initialize the sampling sequences, namely population.}
	\SetKwProg{Fn}{Function}{:}{}
	\Fn{\FRandom{$pop$, $L_g$}}{ %\tcp*{this is another comment}
		$g$ = $randomGenerator(L_g)$\; \tcp*{randomly generate genes $g$ by $L_g$}
		$s$ = $validate(g)$\; \tcp*{obtain evaluation scores $s$ by the initial genes}
		$pop$ = $update(pop, g, s)$ \; \tcp*{update the genes queue with scores}
		\KwRet $pop$\;
	}
	% Write Function with word ``Function''
	\tcc{Select parents from the population and crossover their genes to generate the children's genes}
	\SetKwProg{Fn}{Function}{:}{}
	\Fn{\FCrossover{$pop$, $p_c$, $L_g$}}{
		$father, mother = popout(pop)$\; \tcp*{randomly pop out two elements of queue as parents}
		\For{$i=0, i<L_g, i++$} { \tcp*{go over the entire gene sequence and randomly select the genes from the parents}
			$p = Random(0, 1) $\\
			\If{$p > p_c$}
			{
				$child[i] = father[i]  $
			}
			\Else
			{
				$child[i] = mother[i]  $
			}
		}
		$s_c$ = $validate(child)$\;
		$pop$ = $update(pop, child, s_c)$ \;  \tcp*{if new genes from child is better, update the population.}
		\KwRet $pop$\;
	}
	
	% Write Function with word ``Def''
	\tcc{Randomly select a population and mutate its genes}
	\SetKwProg{Fn}{Function}{:}{}
	\Fn{\FMutation{$pop$, $p_m$, $L_g$}}{
		$g = popout(pop)$\; 
		\For{$i=0, i<L_g, i++$} { 
			$p = Random(0, 1) $\\
			\If{$p > p_m$}
			{
				$g[i] = mutate[i] $ \tcp*{mutate a gene}
			}
			
		}
		$s_m$ = $validate(g)$\;
		$pop$ = $update(pop, g, s_m)$ \; 
		\KwRet $pop$\;
	}
	\tcc{Main function. Loop mutation and crossover operation to generate different offsprings}
	\SetKwProg{Fn}{Function}{:}{\KwRet}
	\Fn{\FMain}{
		$initialize(pop, L_g, p_c, p_m, epoch)$ \;\tcp*{initialize the hyper-parameters}
		$pop$ = Random($pop$, $L_g$)\;
		\For{$i=0, i<epoch, i++$} { 
			$pop$ = Mutation($pop$, $L_g$, $p_m$) \;
			$pop$ = Crossover($pop$, $L_g$, $p_c$)
		}
		best = $popout(pop)$\;
		\KwRet 0\;
	}
	\caption{Evolutionary Algorithm for Non-uniform Sampling}
	\label{evolutionary}
\end{algorithm}

\begin{figure}[t]
	%\vspace{-0.5cm}
	\centering
	\includegraphics[width=0.45\textwidth]{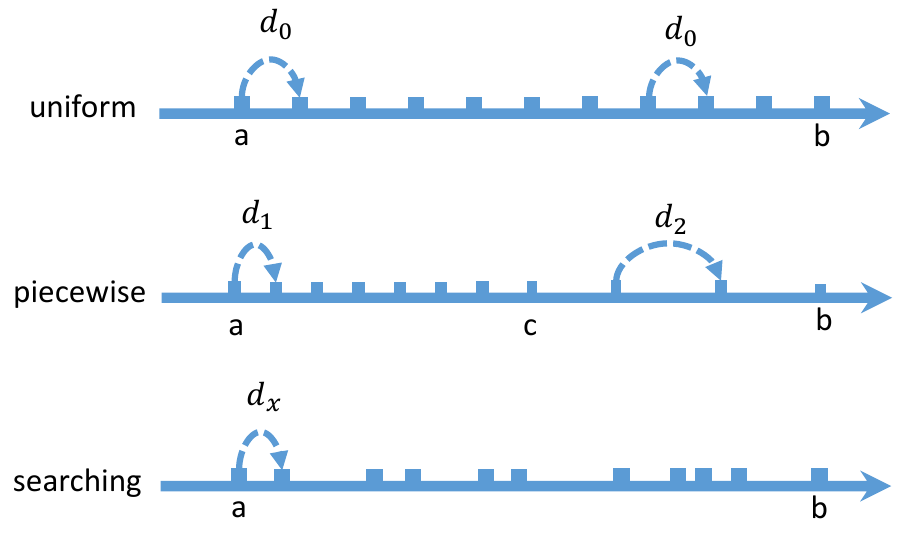}
	%\vspace{-0.4cm}
	\caption{Different sampling strategies. From top to bottom, there are traditional uniform sampling, the proposed piecewise sampling and the searching sampling.}
	\label{sampling_fig}
	%\vspace{-5px}
\end{figure}

\begin{table*}[t]
	\centering
	\caption{Quantitative comparison on the LSUI and UIEB datasets.}
	%\vspace{-0.2cm}
	\label{comparison}
	\begin{tabular}{c|cc|cc|cc}
		\hline
		\multirow{2}{*}{Method} & \multirow{2}{*}{Param.} & \multirow{2}{*}{Time} & \multicolumn{2}{c|}{LSUI}         & \multicolumn{2}{c}{UIEB}          \\
		&                         &                       & PSNR $\uparrow$ & SSIM $\uparrow$ & PSNR $\uparrow$ & SSIM $\uparrow$ \\ \hline
		Fusion~\cite{ancuti2012enhancing}                  & -                       & 1.23s                 & 17.69           & 0.6441          & 18.79           & 0.7921          \\
		MMLE~\cite{MMLE}                  & -                       & 0.30s                 & 17.70           & 0.7252          & 19.30           & 0.8304          \\
		HLRP~\cite{zhuang2022underwater}                  & -                       & 0.32s                 & 12.64 & 0.1929 & 12.56 & 0.2514          \\
		TACL~\cite{liu2022twin}                   & 11M    & 0.01s & 20.69 & 0.8228 & 23.09 & 0.8838          \\
		WaterNet~\cite{li2019underwater}               & 25M                     & 0.55s                 & 22.99           & 0.7890          & 20.48           & 0.7892          \\
		FUnIE~\cite{islam2020fast}                  & 7M                      & 0.02s                 & 18.78           & 0.6196          & 17.61           & 0.5956          \\
		UGAN~\cite{fabbri2018enhancing}                   & 57M                     & 0.06s                 & 22.79           & 0.7545          & 20.59           & 0.6821          \\
		UIE-DAL~\cite{uplavikar2019all}                & 19M                     & 0.04s                 & 21.12           & 0.7231          & 17.00           & 0.7553          \\
		Ucolor~\cite{li2021underwater}                 & 157M                    & 1.87s                 & 22.91           & 0.8902          & 20.78           & 0.8721          \\
		Ushape~\cite{peng2021u}                  & 66M                     & 0.04s                 & 24.16           & \textbf{0.9322}          & 22.91           & 0.9100          \\ 
		
		% AutoEnhancer~\cite{Tang_2022_ACCV}                  & 12M                     & 0.03s                 & 26.13           & 0.8608          & 25.45           & 0.9231          \\ 
		
		\hline
		Ours                    & 10M                     & 0.13s                 & \textbf{27.65}           & 0.8867          & \textbf{28.20}              & \textbf{0.9429}                 \\ \hline
	\end{tabular}
\vspace{0.2cm}
\end{table*}

Note the $Main(\cdot)$ function in the pseudocode, the algorithm includes $epoch$ iterations. In each iteration, the mutation and crossover process are sequentially executed to generate new sampling sequences and validate their performance. In the beginning, we will generate descending and non-repeating sequences as the initial genes. However, with the random mutation or crossover, the offspring sequence may appear repeating elements or turn to non-descending. Therefore, in the implementation, the $validate(\cdot)$ function will be firstly used to remove the illegal offspring sequences and then execute the inference process of our diffusion model by these legal sequences to obtain scores. Here, we use PSNR to rank the scores of the top K sampling sequence and update the queue by the $update(\cdot)$ function. Based on this algorithm, we can obtain the optimal sampling sequence for the inference of the diffusion model against the traditional uniform sampling methods.

In summary, Figure.~\ref{sampling_fig} shows the difference between the uniform sampling strategy and the proposed two non-uniform sampling strategies. The coordinate axis denotes the sampling sequence in interval $[a, b]$. The points in the axis are the sampling values, namely $\tau_s$. Firstly, the traditional uniform sampling adopts the same time interval $d_0$. Secondly, we propose piecewise sampling, whose intervals have two values. The sampling intervals are $d_1$ and $d_2$ in the $[a, c]$ and $[c, b]$, respectively. Finally, we propose to sample the sequence by using the searching algorithm. The intervals are arbitrary and obtained by the searching, so they are represented by $d_x$.

\section{Experiments}

\subsection{Experimental settings}
\textbf{Datasets.} As well known, the enhancement ground truths for underwater images are hard to obtain. Previous datasets like RUIE\cite{liu2020real}, U-45 \cite{li2019fusion} do not provide the paired data, namely underwater images and their corresponding ground truths. These datasets are inconvenient for network training and testing. In this paper, two recently released datasets are used for network training and evaluation. 

\textbf{Underwater Image Enhancement Benchmark (UIEB)}~\cite{li2019underwater}. This dataset contains 890 paired images. The underwater images are captured from the Internet and the ground truths are generated by the previous enhancement methods and manual selection. Concretely, several enhancement approaches are used to generate their corresponding enhanced results by giving the collected underwater images. Then, the best result is chosen by several volunteers' voting and treated as the ground truth. In this paper, following the previous setting in~\cite{li2019underwater}, we adopt the testing set, containing 90 images, to evaluate our approach.

\begin{figure*}[t]
	%\vspace{-0.2cm}
	\centering
	\includegraphics[width=1\textwidth]{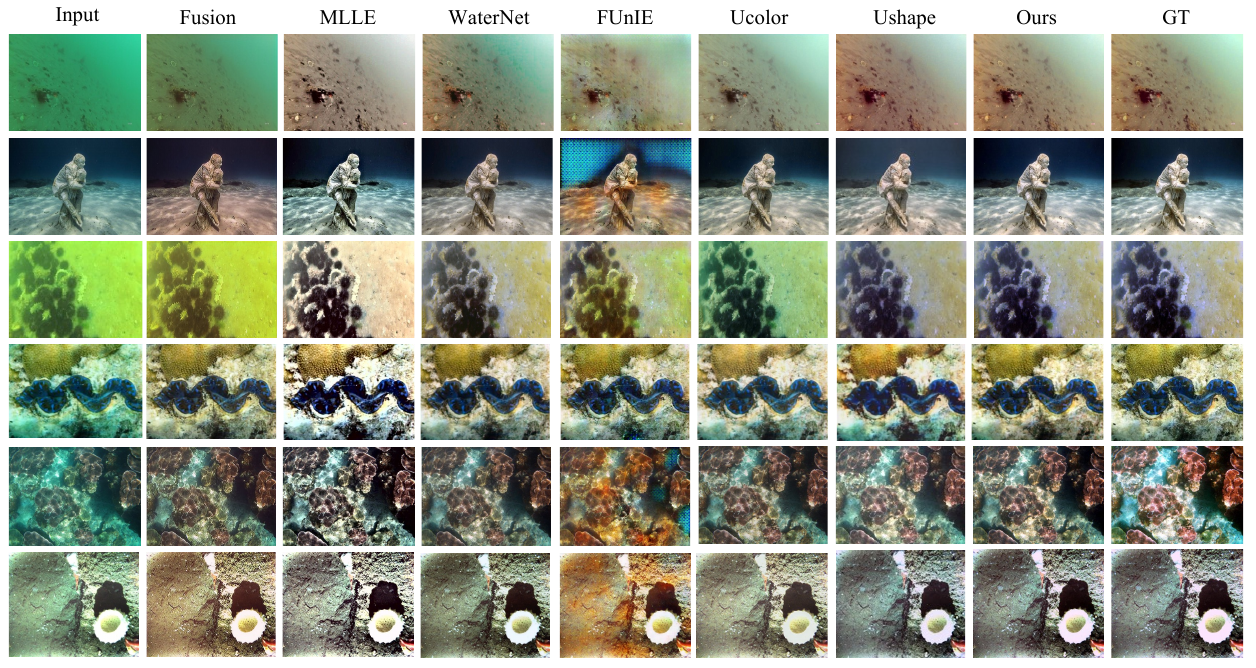}
	%\vspace{-0.8cm}
	\caption{Visual comparison on the underwater dataset. The underwater images and their corresponding enhanced images from previous methods and the proposed one are presented. Besides, the ground truths are displayed in the last columns. }
	\label{show}
	\vspace{0.3cm}
\end{figure*}

\textbf{Large-Scale Underwater Image (LSUI)}~\cite{peng2021u}. Considering the shortage of training images in the previous datasets, LSUI follows the labeling methods in~\cite{li2019underwater} and collects a larger dataset. Compared with the previous one, LSUI dataset contains 5004 underwater images and their corresponding high-quality images. Moreover, LSUI contains diverse underwater scenes, object categories and deep-sea and cave images. In this paper, we use the training set of LSUI, namely 4500 paired images to train the diffusion model. The remaining 504 images are to validate the proposed approach. 

%The collection of this dataset almost follows the rule of UIEB, but LSUI is much larger than UIEB. In order to satisfy the training requirements, LSUI collects 5004 underwater images and their corresponding high-quality images. For the setting in~\cite{peng2021u}, 4500 paired images are used for training. The remaining 504 images are used for testing.

\textbf{Evaluation Metrics.} Previous methods usually employ subjective evaluation standards, such as UCIQE~\cite{yang2015underwater} and UIQM~\cite{panetta2015human}. However, according to the statement in~\cite{peng2021u}, these two metrics cannot accurately measure the approaches in some scenes. In this paper, we mainly exploit two full reference evaluation measures: Peak Signal to Noise Ratio (PSNR) and Structure SIMilarity index (SSIM). Both of them represent the closeness to the reference, where the PSNR value reflects how close to the image content and the SSIM value reflects how similar to structure and texture.

\subsection{Implementation details}

In this paper, we adopt PyTorch to implement the proposed approach. For our network training, we employ the Adam optimizer to minimize the objective function. The learning rate is set to 1.0 $\times$10$^{-4}$. During the training stage, to balance the batch size and image size, the batch and image size are set to 8 and 128 $\times$ 128, respectively. The pixel values of the image are normalized to [-1,1]. The time step of the diffusion model is set to 2000. the $\beta$ is linearly sampled within the range $[10^{-6}, 10^{-2}]$. In the testing stage, following the setting in~\cite{peng2021u}, the input size of the image is 256 $\times$ 256. By using the skip sampling strategy, the sampling times are set to 10 times to balance the performance and runtime. The hardware we use for training and testing is a workstation with an NVIDIA RTX 3080. The entire training time is about 35 hours with our workstation.  

\begin{figure*}[ht]
	%\vspace{-0.5cm}
	\centering
	\includegraphics[width=0.8\textwidth]{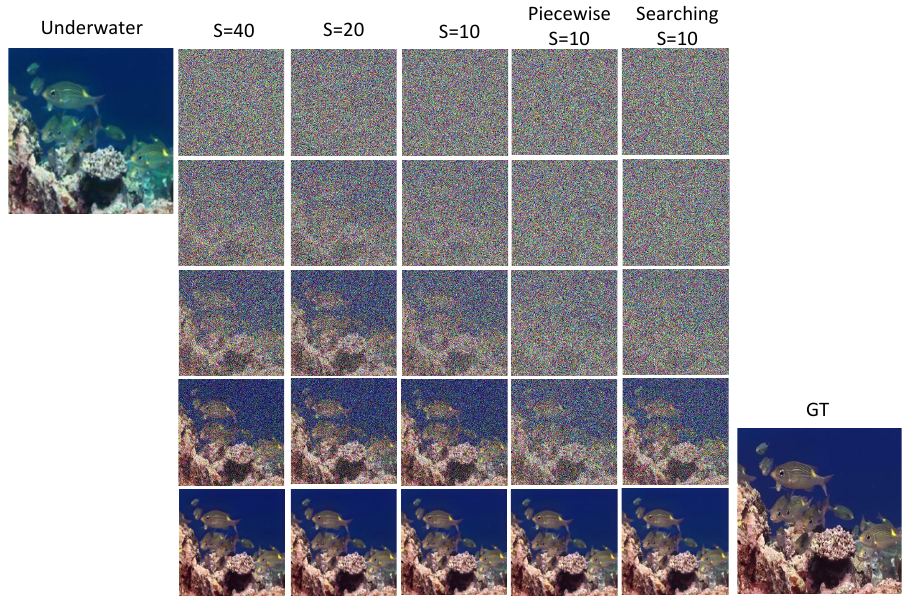}
	%\vspace{-0.4cm}
	\caption{Visual comparison with different time steps. The left top is the underwater image and the bottom right is the corresponding ground truth. We validate 4 different numbers of iterations and the proposed two sequence sampling methods.}
	\label{show_s}
	\vspace{0.2cm}
\end{figure*}

\subsection{Comparison with the state-of-the-arts}

In this paper, we compare with eight state-of-the-art techniques, such as Fusion~\cite{ancuti2012enhancing}, MMLE~\cite{MMLE}, HLRP~\cite{zhuang2022underwater}, TACL~\cite{liu2022twin}, WaterNet~\cite{li2019underwater}, FUnIE~\cite{islam2020fast}, UGAN~\cite{fabbri2018enhancing}, UIE-DAL~\cite{uplavikar2019all}, Ucolor~\cite{li2021underwater}, and Ushape~\cite{peng2021u}. Among them, Fusion~\cite{ancuti2012enhancing}, MMLE~\cite{MMLE} and HLRP~\cite{zhuang2022underwater} are traditional non-deep learning-based methods. The remaining approaches are deep learning models. 

Results are summarized in Table.\ref{comparison}, where we can confirm that the proposed method achieves the best performance on PSNR and competitive quality on SSIM. Notice that compared with Ushape~\cite{peng2021u}, the proposed approach obtain performance gains of 3.49 dB, which is a significant improvement. Moreover, due to the employment of lightweight transformer blocks, our deep denoising network is much smaller than that of Ushape. Although the number of parameters in FUnIE~\cite{islam2020fast} is slightly smaller than ours, their performance is not sufficient against the others. However, the value of our SSIM cannot achieve the state-of-the-art.  We know that SSIM means structural similarity, which needs to compare luminance, contrast and structure. One reason is that our diffusion model tries to recover the clean image by random Gaussian noise from the beginning. Some structural information such as edge was harder to recover than the color information, which might decrease the perceptual value of SSIM, whereas Ushape~\cite{peng2021u} is based on a simple encoder-decoder model, which can keep such a structure. 

Based on those findings, it is confirmed that the proposed method can greatly balance the model size and performance, and achieve competitive performance on both sides. Besides, the runtime of the diffusion network is just slightly slower than previous methods~\cite{peng2021u,uplavikar2019all}. The reason is that the diffusion-based network needs to process an iterative process like the recurrent network. In this paper, due to the introduction of the skip sampling strategy, the process can be remarkably accelerated. The runtime of our model can achieve 0.13s for each image. The choice of the hyperparameters and validation experiments are displayed later. 

Figure.\ref{show} presents the visual comparison between the proposed methods and the other state-of-the-art approaches. Notice that the ability of color correction by our approach is outstanding. For example, Ucolor can remove the green noise in the underwater image to some extent, but its enhanced image cannot restore the original object's color. A similar phenomenon also appears in the enhanced images by Ushape. The reddish color is generated by Ushape in the enhanced image of the second row, whereas no such effect in ours. From the visual inspection compared with the previous methods, it is confirmed that the proposed method has achieved competitive performance in color correction and restoration. 
%Moreover, the robustness of our method is also remarkable. It can remove the different types of noise in the different underwater scenes.

%the green noise in the first row and the blue noise in the second row are completely removed by our model. 

\subsection{Ablation studies}

\begin{table}[]
	\centering
	\caption{The comparison of different denoising networks on LSUI dataset.}
	%\vspace{-0.5cm}
	\label{compare_t}
	\begin{tabular}{|c|c|c|c|c|}
		\hline
		Denoising network & Param. & Time  & PSNR  & SSIM   \\ \hline
		UNet              & 32M    & 0.13s & 24.52 & 0.8671 \\ \hline
		Ours              & 10M    & 0.13s & 27.65 & 0.8867 \\ \hline
	\end{tabular}
	%\end{table}
	\vspace{0.2cm}
	%\begin{table}[]
	\centering
	\caption{The comparison of different skip values on the LSUI dataset. $S$ represents the number of iterations by using a denoising network in the reverse process.}
	%\vspace{-0.2cm}
	\label{compare_s}
	\begin{tabular}{|c|c|c|c|}
		\hline
		S (times) & Runtime(s) & PSNR  & SSIM\\ \hline
		40        & 0.73    & 27.64 & 0.8857 \\ \hline
		20        & 0.37    & 27.62 & 0.8842 \\ \hline
		10        & 0.13    & 27.57 & 0.8851 \\ \hline
		$Ours_{p}$ & 0.13    & 27.61 & 0.8859 \\ \hline
		$Ours_{s}$ & 0.13    & 27.65 & 0.8867 \\ \hline
	\end{tabular}
	%\vspace{-0.5cm}
\end{table}

In this paper, we propose a Transformer-based architecture as the denoising neural network in the diffusion framework. In this section, we firstly validate the effectiveness of the proposed network against the widely used network, namely UNet whose architecture is the traditional convolution-based structure with gradually increasing channel numbers in the encoder and decreasing channel numbers in the decoder. The experimental results are shown in Table.\ref{compare_t}. Notice that the proposed Transformer-based architecture can obtain great improvement against the traditional UNet. Besides, the model size of the proposed one is smaller, which makes the runtime faster. 

Next, the other experiment is to find a suitable skip time interval by using~\cite{song2020denoising}. Generally, previous methods employ uniform distribution to sample the time step. For example, when the training time step is set to 2000 (namely $T=2000$) and the sampling step in the reverse process is set to 20 (namely $S=20$), the time interval is 100. Then, the sampled time sequence will be $\{2000, 1800, ..., 200, 0\}$, which is treated as the input to feed into the denoising network together with noise image $x_t$ and conditional image $c$. In our experiment, to find a suitable trade-off, we try to use different numbers of iterations to restore the images in the reverse process. We choose $S = \{40, 20, 10\}$ and their results are shown in Table.\ref{compare_s}. We can confirm that the values of PSNR and SSIM keep high for all the cases, whereas, the runtime is decreasing dramatically when the $S$ turns small. %Therefore, a suitable trade-off is $S=10$, which can balance the performance and runtime. After that, 

Finally, we keep the number of iterations, namely $S=10$, and validate the effectiveness of the proposed sampling approaches. $Our s_p$ represents the proposed piecewise sampling strategy and $Our s_s$ represents the searching strategy by the evolutionary algorithm, where the optimal interval is automatically decided. Results are shown in Table.\ref{compare_s}, fourth and fifth row. From the results, we can confirm that both approaches can improve the performance, however, $Ours_s$ is slightly better than $Our s_p$, where  $Ours_s$ reaches the PSNR of $S= 40$ and surpasses the value of $S= 20$. Figure.\ref{show_s} presents the visual comparison by using the different numbers of iterations and the proposed sequence sampling methods. It also shows the gradual denoising process from the noise images to the clear images.   

\section{Conclusion}

In this paper, we introduce a conditional diffusion model for underwater image enhancement. In our framework, the network is able to generate the corresponding enhanced image by giving the underwater one. Moreover, the generated results achieve competitive performance against the state-of-the-art methods in recent years. It is achieved by our lightweight transformer-based denoising network to remove Gaussian noises from the generated images to make clear ones. The proposed network not only achieves remarkable performance but also improves the efficiency of each network forward in the reverse process, thus speeding up the entire inference process of the diffusion model. Finally, we propose two different sampling approaches for the sequence of the time step based on the skip sampling strategy, namely piecewise and searching sampling with the evolutionary algorithm. Both of them can further improve the performance by using the same number of time steps against the previous uniform sampling. Our future work tries to extend the technique to an unsupervised one without using synthetic images.

\bibliographystyle{ACM-Reference-Format}
\bibliography{sample-base}


\end{document}